\documentclass[10pt,twocolumn,letterpaper]{article}

\usepackage{cvpr}
\usepackage{times}
\usepackage{epsfig}
\usepackage{graphicx}
\usepackage{amsmath}
\usepackage{amssymb}

\usepackage{subfigure}

\usepackage{algorithm}
\usepackage{algorithmic}

\usepackage[breaklinks=true,colorlinks,bookmarks=false]{hyperref}

\cvprfinalcopy 

\def\cvprPaperID{****} 

\begin{document}


\title{ContourNet: Taking a Further Step toward Accurate Arbitrary-shaped Scene Text Detection}

\author{Yuxin Wang, Hongtao Xie\thanks{Corresponding author} , Zhengjun Zha, Mengting Xing, Zilong Fu and Yongdong Zhang\\
University of Science and Technology of China\\
{\tt\small \{wangyx58,metingx,JeromeF\}@mail.ustc.edu.cn,\{htxie,zhazj,zhyd73\}@ustc.edu.cn}
}

\maketitle
\thispagestyle{empty}
\pagestyle{empty}

\begin{abstract}
Scene text detection has witnessed rapid development in recent years. However, there still exists two main challenges: 1) many methods suffer from false positives in their text representations; 2) the large scale variance of scene texts makes it hard for network to learn samples. In this paper, we propose the ContourNet, which effectively handles these two problems taking a further step toward accurate arbitrary-shaped text detection. At first, a scale-insensitive Adaptive Region Proposal Network (Adaptive-RPN) is proposed to generate text proposals  by only focusing on the Intersection over Union (IoU) values between predicted and ground-truth bounding boxes. Then a novel Local Orthogonal Texture-aware Module (LOTM) models the local texture information of proposal features in two orthogonal directions and represents text region with a set of contour points. Considering that the strong unidirectional or weakly orthogonal activation is usually caused by the monotonous texture characteristic of false-positive patterns (\emph{\eg} streaks.), our method effectively suppresses these false positives by only outputting predictions with high response value in both orthogonal directions. This gives more accurate description of text regions. Extensive experiments on three challenging datasets (Total-Text, CTW1500 and ICDAR2015) verify that our method achieves the state-of-the-art performance. Code is available at \url{https://github.com/wangyuxin87/ContourNet}.
\end{abstract}

\section{Introduction}

Scene text detection is a task to detect text regions in the complex background and label them with bounding boxes. Accurate detection result benefits a wide scope of real-world applications and is the fundamental step for end-to-end text recognition \cite{wang2019dsrn,fang2018attention,xie2019convolutional,lyu2018mask}.

Benefiting from the development of deep learning, recent methods have achieved significant improvement in scene text detection task. Meanwhile, the research focus has shifted from horizontal texts \cite{zhang2015symmetry,liao2017textboxes} to multi-oriented  texts \cite{LyuYWYB18,ZhouYWWZHL17} and more challenging arbitrary-shaped texts \cite{wang2019efficient,wang2019arbitrary} (\eg curved texts). However, due to the specific properties of scene text such as large variance in color, texture, scale, etc., there are still two challenges to be addressed in arbitrary-shaped scene text detection.

\begin{figure}[t!p]
\label{model}
  \centering
  \includegraphics[width=8cm]{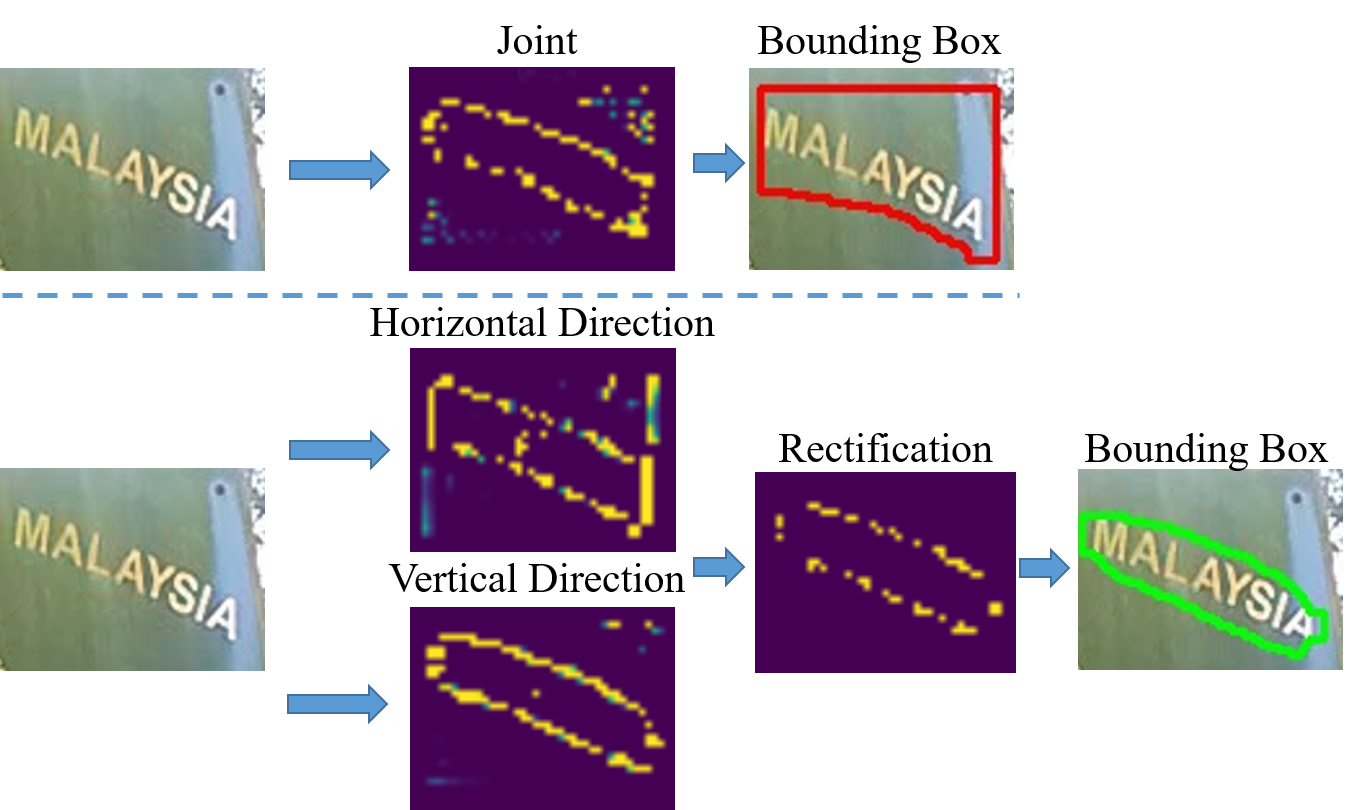}\\
  \caption{ Comparison between jointly modeling the texture information in arbitrary orientations and respectively modeling texture information in two orthogonal directions. We visualize the heatmap of predicted contour points. NMS is implemented to each heatmap for better visualization. FPs can effectively be suppressed by considering the response in two orthogonal directions simultaneously.
}\label{recent}
 \vspace{-1em}
\end{figure}

The first challenge is false positives (FPs), which is not received enough attention in recent researches \cite{xie2019scene} and is regarded as one of the key bottlenecks for more accurate arbitrary-shaped scene text detection in this paper. Recent CNN-based methods jointly model the texture information in arbitrary orientations by using $k \times k$ convolutional kernel \cite{zhang2019look,xue2019msr}. However, this operation is sensitive to some specific cases containing similar texture characteristics to text regions, and tends to perform identically high response to these cases (see in top of Fig.\ref{recent}). SPCNET \cite{xie2019scene} attributes this problem to the lack of context information clues and inaccurate classification scores, thus a text context module is proposed to compensate global semantic feature and bounding boxes are further rectified by the segmentation map. Liu \emph{et al.} \cite{liu2019omnidirectional} re-score the detection results with the confidence of four vertexes to supervise the compactness of the bounding boxes. Different from these methods, we handle the FPs by using only local texture information, which is a more straightforward approach and contains less computation. As shown in bottom of Fig.\ref{recent}, our motivation mainly comes from two observations: 1) FPs with strong unidirectional texture characteristics are weakly activated in its orthogonal direction (\eg some vertical streaks); 2) FPs can be effectively suppressed by considering the responses in both orthogonal directions simultaneously. Thus, it is reasonable to model the texture information along two orthogonal directions. Inspired by traditional edge detection operators (\eg Sobel, etc.), we heuristically use horizontal and vertical directions in our approach.

The second challenge is the large scale variance of scene texts. Compared with general objects, the scale variation is much larger in scene texts, which makes it hard for CNN-based methods to learn samples. To address this problem, MSR \cite{xue2019msr} uses a multi-scale network to obtian powerful representation of texts with various scales. DSRN \cite{wang2019dsrn} attributes this problem to the inconsistent activation of multi-scale texts, thus a bi-directional operation is proposed to map the convolutional features to a scale-invariant space. Different from these methods solving the large scale variance problem through aggregation of multi-scale features, we pay attention to the shape information and use a scale-invariant metric to optimize our network.

\begin{figure*}[t!p]
  \centering
  \includegraphics[width=16.5cm]{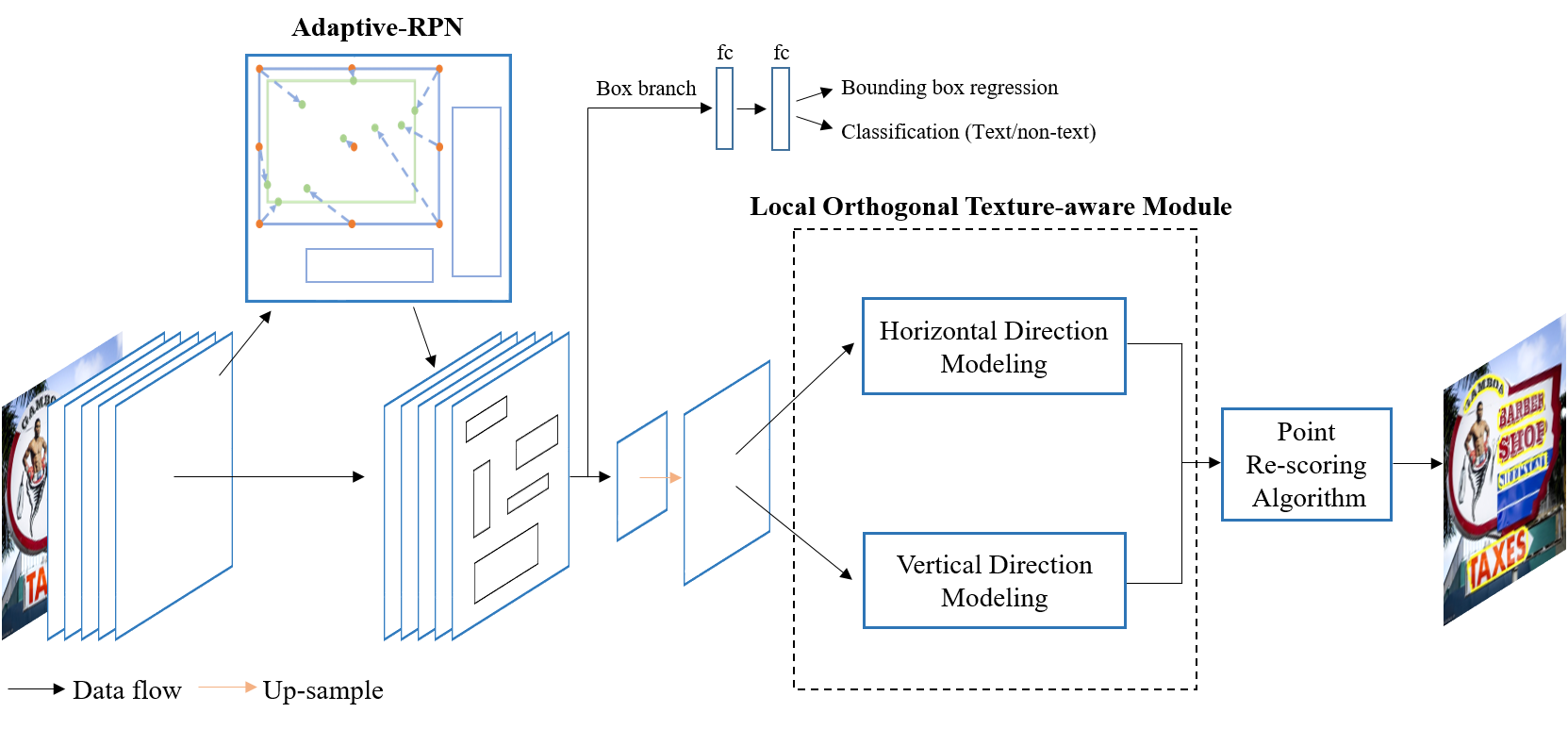}\\
  \vspace{-1em}
  \caption{ The pipeline of ContourNet. It mainly contains three parts: Adaptive Region Proposal Network (Adaptive-RPN), Local Orthogonal Texture-aware Module (LOTM) and \emph{Point Re-scoring Algorithm}. The box branch is similar to other 2-stage methods.}\label{f111}
  \label{model_total}
  \vspace{-1em}
\end{figure*}

In this paper, we propose a novel text detector to effectively solve these two problems achieving accurate arbitrary-shaped scene text detection, which is called ContourNet. As shown in Fig.\ref{model_total}, given an input image, \emph{Adaptive Region Proposal Network} (Adaptive-RPN) first generates text proposals by automatically learning a set of boundary points over the text region that indicate the spatial extend of text instance. The training object of Adaptive-RPN is driven by \emph{IoU} values between the predicted and ground-truth bounding boxes, which is invariant to the scale \cite{rezatofighi2019generalized,ZhouYWWZHL17}. Thus, Adaptive-RPN is insensitive to the large scale variance of scene texts and can automatically account for shape information of text regions to achieve finer localization compared with conventional RPN approaches \cite{ren2015faster,he2017mask}. To capture the distinct texture characteristics in contour regions of texts, we propose a \emph{Local Orthogonal Texture-aware Module} (LOTM) to model the local texture information of proposal features in two orthogonal directions, and represent text region with contour points in two different heatmaps, either of which only responds to the texture characteristics in a certain direction. Finally, \emph{Point Re-scoring Algorithm} effectively filters predictions with strong unidirectional or weakly orthogonal activation by considering the response in both orthogonal directions simultaneously. In this way, text regions are detected and represented with a set of high-quality contour points.

The contributions of this paper are three-fold: 1) We propose a novel approach for FP suppression by modeling the local texture information in two orthogonal directions, which is a more straightforward approach and contains less computation compared with previous methods. 2) The proposed Adaptive-RPN effectively handles the large scale variance problem and achieves finer localization of text regions, which can be easily embedded into existing approaches. 3) Without external data for training, the proposed method achieves \textbf{85.4\%} and \textbf{83.9\%} in F-measure on Total-Text and CTW1500 dataset with 3.8 FPS and 4.5 FPS respectively, which outperforms recent counterparts by a large margin.

\section{Related Works}

Scene text detection has been a popular research topic for a long time with many approaches proposed \cite{zhang2015symmetry,tian2015text,yin2013robust,zhang2019look,LyuYWYB18,xue2018accurate,wang2019efficient,tian2019learning,Wang_2019_CVPR}. Conventionally, connected component (CC) based and sliding window based methods have been widely used in text localization \cite{zhang2015symmetry,tian2015text,yin2013robust}. As deep learning becomes the most promising machine learning tool \cite{xu2018dual,liu2016hierarchical,8907499,zhang2012attribute}, scene text detection has achieved remarkable improvement in recent years. These methods can be divided into two categories: regression based methods and segmentation based methods.

\textbf{Regression based methods} \cite{tian2017wetext,ZhouYWWZHL17}, inspired by general object detection methods \cite{girshick2015fast,DBLP:conf/eccv/LiuAESRFB16,he2017mask}, localize text boxes by predicting the offsets from anchors or pixels. Lyu \emph{et al.} \cite{LyuYWYB18} adopt a similar architecture as SSD and rebuild text instance with predicted corner points. Wang \emph{et al.} \cite{wang2019arbitrary} use recurrent neural network (RNN) for text region refinement and adaptively predict several pairs of points to represent arbitrary-shaped text. Different from these methods localizing text regions by implementing refinement on pre-defined anchors, EAST \cite{ZhouYWWZHL17} and DDR \cite{HeZYL17} propose a new approach for accurate and efficient text detection, which directly regress the offsets from boundaries or vertexes to current point. Based on these direct regression methods, LOMO \cite{zhang2019look} proposes an iterative refinement module to iteratively refine bounding box proposals for extremely long texts, and then predicts center line, text region, and border offsets to rebuild text instance.

\textbf{Segmentation based methods} \cite{DBLP:conf/eccv/LongRZHWY18,wang2019efficient} are mainly inspired by FCN \cite{DBLP:conf/cvpr/LongSD15}. Recent segmentation based methods usually use different representation to describe text region, and then rebuild text instance through specific post-procession. PixelLink \cite{DBLP:conf/aaai/DengLLC18} predicts connections between pixels and localizes text region by separating the links belonging to different text instances. To handle the adjacent texts, Tian \emph{et al.} \cite{tian2019learning} design a two-step clustering to split dense text instances from segmentation map. PSENet \cite{wang2019efficient} gradually expands kernels at certain scale to split the close text instances.

Our method integrates the advantages of regression based methods and segmentation based methods, which adopts a two-stage architecture and represents text region with contour points. Benefiting from Adaptive-RPN and FP suppression, our method effectively handles the large scale variance problem and gives more accurate description of text regions compared with previous methods.

\section{Proposed Method}

The proposed method mainly consists of three parts: Adaptive-RPN, LOTM and \emph{Point Re-scoring Algorithm}. In this section, we first briefly describe the overall pipeline of the proposed method, and then detail the motivation and implementation of these three parts respectively.

\subsection{Overall pipeline}

The architecture of our ContourNet is illustrated in Fig.\ref{model_total}. First, a backbone network is constructed to generate shared feature maps. Inspired by FPN \cite{LinDGHHB17} which can obtain strong semantic features for multi-scale targets, we construct a backbone with FPN-like architecture by implementing lateral connections in the decoding layer. Next, we propose an Adaptive-RPN described in Sec.3.2 for proposal generation by bounding spatial extent of several refined points. The input of LOTM are proposal features obtained by using Deformable RoI pooling \cite{zhu2019deformable} and bilinear interpolation to the shared feature maps. Then, LOTM decodes the contour points from proposal features by modeling the local texture information in horizontal and vertical directions respectively. Finally, a \emph{Point Re-scoring Algorithm} is used to filter FPs by considering the responses in both directions simultaneously. The details of LOTM and \emph{Point Re-scoring Algorithm} are presented at Sec.3.3 and 3.4 respectively. Bounding box regression and classification (text/non-text) in box branch are similar to other 2-stage methods, which are used to further refine bounding boxes.

\subsection{Adaptive Region Proposal Network}

Region Proposal Network is wildly used in existing object detection methods. It aims to predict a 4-d regression vector $\{\Delta x, \Delta y, \Delta w, \Delta h\}$ to refine current bounding box proposal $B_c = \{x_c, y_c, w_c, h_c\}$ to a predicted bounding box $B_t = \{x_c + w_c \Delta x_c, y_c + h_c \Delta y_c, w_ce^{\Delta w_c}, h_ce^{\Delta h_c}\}$, and the training objective is to optimize the smooth ${l_1}$ loss \cite{ren2015faster}.

As an approach proposed to improve the \emph{IoU} value between predicted and ground-truth bounding boxes, this aforementioned $4-d$ representation optimized by ${l_n}$-norm loss is sensitive to the scale variation. In general, positive bounding boxes are selected through an \emph{IoU} metric (\eg $IoU > 0.5$). However, several pairs of bounding boxes in different scales with the same \emph{IoU} value may have different ${l_n}$-norm distances. As there is not a powerful correlation between optimizing ${l_n}$-norm loss and improving their \emph{IoU} values \cite{rezatofighi2019generalized}, we infer that this gap makes it hard for CNN-based methods to learn samples with large scale variance in scene text detection.

To handle this problem, we propose a new Adaptive-RPN to focus on only \emph{IoU} values between predicted and ground-truth bounding boxes which is a scale-invariant metric, and use a set of pre-defined points $P=\{ (x_l, y_l) \}_{l=1}^n$ (1 center point and $n-1$ boundary points) instead of the 4-d vector for the proposal representation. The refinement can be expressed as:
\vspace{-1em}

\begin{equation}\label{regress_cp}
{
R=\{x_r, y_r \}_{r=1}^n = \{ (x_l + w_c \Delta x_l, y_l + h_c \Delta y_l) \}_{l=1}^n
}
\end{equation}

\begin{figure}[t!p]
  \centering
  \includegraphics[width=7cm]{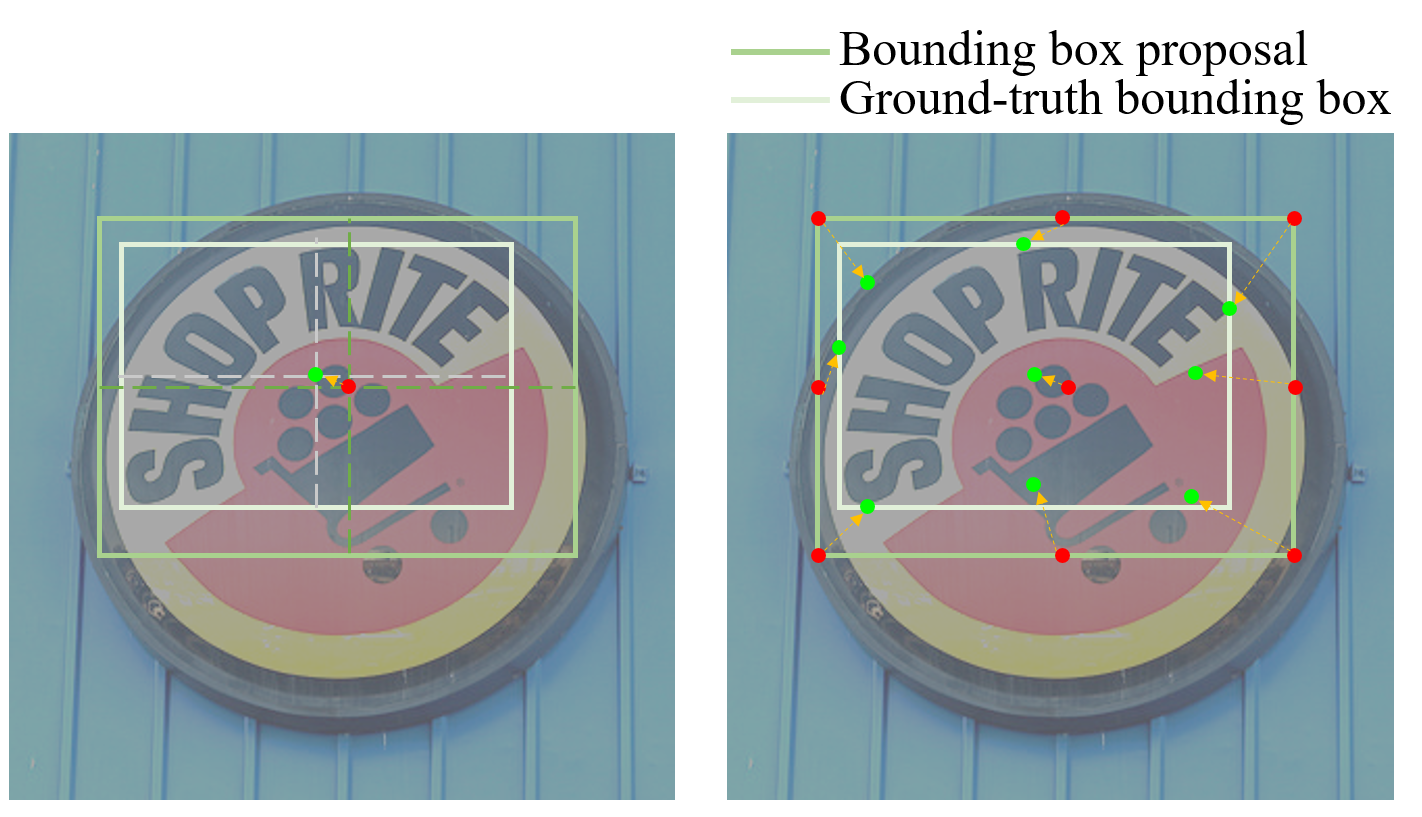}\\
  \caption{The comparison between conventional RPN (left) and Adaptive-RPN (right). The proposed Adaptive-RPN adaptively regresses the offsets to pre-defined points. Predicted bounding box is generated by bounding the spatial extend of refined points. Red points are pre-defined points in current bounding box proposal (\eg center point in conventional RPN and pre-defined
points \emph{P} in Adaptive-RPN), and green points are refined points. The yellow dotted lines indicate the regressed offsets.}\label{rpn}
  \vspace{-1em}
\end{figure}

Where $\{\Delta x_l,\Delta y_l\}_{l=1}^n$ are the predicted offsets to pre-defined points, $w_c$ and $h_c$ are width and height of current bounding box proposal. As shown in Fig.\ref{rpn}, the predicted offsets are used to process a local refinement on \emph{n} pre-defined points in current bounding box proposal. Then, we use a max-min function in eq.(\ref{refine}) to bound these refined points with 4 extreme points for the representation of predicted bounding box. Specially, the center point $\{x', y'\}$ is used to normalize the bounding box (\eg if $x_{tl} > x'$, then $x_{tl} = x'$).

\begin{equation}\label{refine}
\begin{split}
Proposal=& \{ x_{tl}, y_{tl}, x_{rb}, y_{rb}\}\\ =& \{ min\{x_r\}_{r=1}^n, min\{y_r\}_{r=1}^n, \\
& max\{x_r\}_{r=1}^n, max\{y_r\}_{r=1}^n\}
\end{split}
\end{equation}

Compared with conventional RPN that considers only rectangular spatial scope, the proposed Adaptive-RPN automatically accounts for shape and semantically important local areas for finer localization of text regions. Without additional supervision, we optimize the regression loss in Adaptive-RPN through an \emph{IoU} loss (see in eq.(\ref{l_rpn})) by calculating the overlapping between the predicted and ground-truth bounding boxes.

\subsection{Local Orthogonal Texture-aware Module}

Inspired by traditional edge detection operators (\eg Sobel, etc.) which have achieved remarkable performance before deep learning becomes the most promising machine learning tool, we skillfully incorporate the idea of traditional edge detection operators into LOTM and represent text region with a set of contour points. These points containing strong texture characteristics can accurately localize texts with arbitrary shapes (rectangular and irregular shapes shown in Fig. \ref{result}).

\begin{figure}[t!p]
  \centering
  \includegraphics[width=8cm]{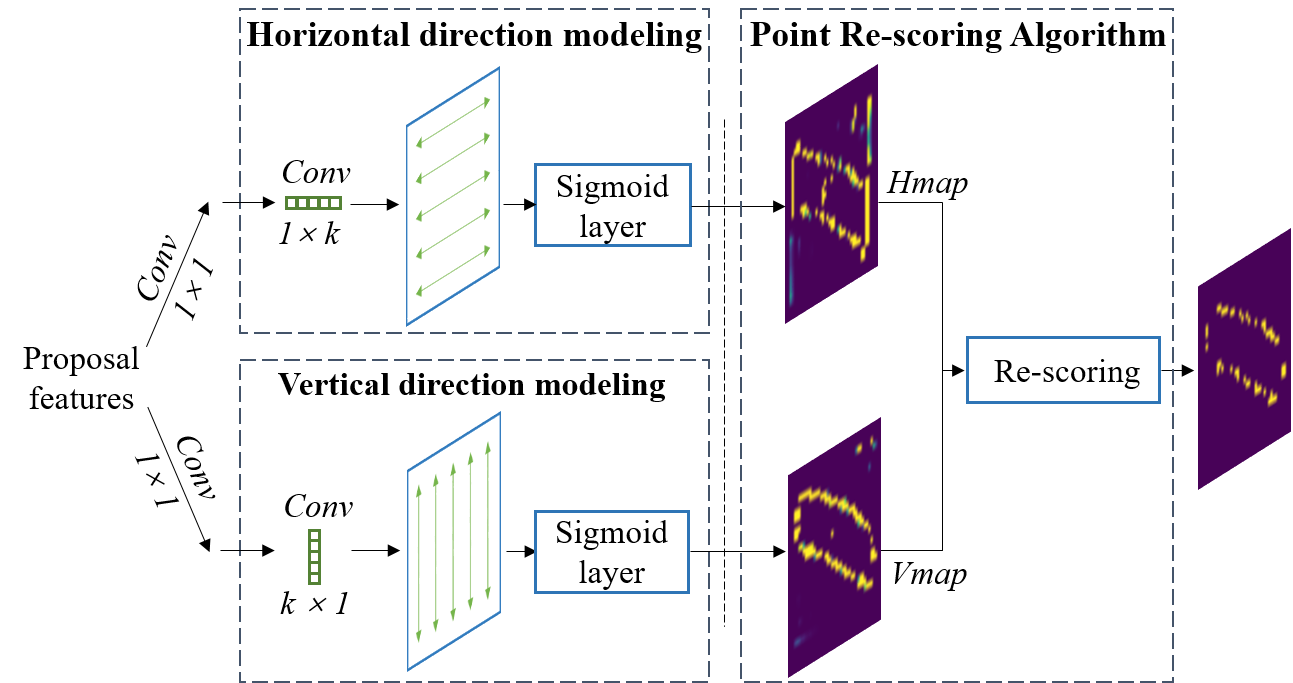}\\
  \caption{The visualization of LOTM (left). \emph{Point Re-scoring Algorithm} (right) is only used in testing stage. }\label{bcdm}
  \vspace{-1em}
\end{figure}

As shown in Fig.\ref{bcdm}, LOTM contains two parallel branches. In the top branch, we slide a convolutional kernel with size $1 \times k$ over the feature maps to model the local texture information in horizontal direction, which only focuses on the texture characteristics in a \emph{k}-range region. This local operation is proved to be powerful in our experiments, and because of the small amount of computation, it also keeps the efficiency of our method. In a like manner, the bottom branch is constructed to model the texture characteristics in vertical direction through a convolutional kernel with size $k \times 1$. \emph{k} is a hyper-parameter to control the size of receptive field of texture characteristics, which is discussed in ablation experiments in Sec.4. Finally, two sigmoid layers are implemented to normalize the heatmaps to $[0,1]$ in both directions. In this way, text regions can be detected in two orthogonal directions and represented with contour points in two differen heatmaps, either of which only responds to texture characteristics in a certain direction.

\subsection{Point Re-scoring Algorithm}

As false-positive predictions can be effectively suppressed by considering the response value in both orthogonal directions, two heatmaps from LOTM are further processed through \emph{Point Re-scoring Algorithm}. As shown in Algorithm \ref{algo1}, points in different heatmaps are first processed through \emph{Non-Maximum Suppression} (NMS) to achieve a tight representation. Then, to suppress the predictions with strong unidirectional or weakly orthogonal response, we only select the points with distinct response in both heatmaps as candidates. Finally, text region can be represented with polygon made up by these high-quality contour points. $NMS_H$ and $NMS_V$ mean NMS operation in horizontal and vertical direction respectively. We set $\theta$ to 0.5 for better trade-off between recall and precision.

\begin{algorithm}[b!p]
	\caption{Point Re-scoring Algorithm}
	\begin{algorithmic}[1]
        \REQUIRE {Heatmaps in two orthogonal  directions: $Hmap$, $Vmap$.}
        \ENSURE {Contour point candidates: $Contourmap$}.
        \STATE $Contourmap=zeros$\_$like(Hmap)$
		\STATE $Hmap=NMS_H(Hmap)$
        \STATE $Vmap=NMS_V(Vmap)$
		\FOR{$(i,j)$ $in$ $Hmap$}
		\IF{$Confidence[Hmap[i,j]] > \theta $}
        \IF{$Confidence[Vmap[i,j]] > \theta $}
		\STATE $Contourmap[i,j]=1$
        \ENDIF
		\ENDIF
		\ENDFOR
        \RETURN $Contourmap$
	\end{algorithmic}
\label{algo1}
\end{algorithm}

\subsection{Training Objective}

For learning ContourNet, the loss function is formulated as:

\begin{equation}\label{l_point}
\begin{split}
L = &L_{Arpn_{cls}} + \lambda_{A_{reg}} L_{Arpn_{reg}} + \lambda_{Hcp} L_{Hcp}  \\
& + \lambda_{Vcp} L_{Vcp} + \lambda_{box_{cls}} L_{box_{cls}} \\
&+ \lambda_{box_{reg}} L_{box_{reg}}
\end{split}
\end{equation}

Where $L_{Arpn_{cls}}$, $L_{Arpn_{reg}}$, $L_{Hcp}$, $L_{Vcp}$, $L_{box_{cls}}$ and $L_{box_{reg}}$ denote Adaptive-RPN classification loss, Adaptive-RPN regression loss, contour point loss in horizontal direction, contour point loss in vertical direction, bounding box classification loss and  bounding box regression loss respectively. We use balance weights $\lambda_{A_{reg}}$, $\lambda_{Hcp}$, $\lambda_{Vcp}$, $\lambda_{box_{cls}}$ and $\lambda_{boxreg}$ to represent the importance among six losses. We simply balance $\lambda_{A_{reg}}$ and set others to 1 in our experiment.

\textbf{Adaptive-RPN:} Adaptive-RPN is optimized with an \emph{IoU} loss to achieve robust performance on scale variance. The loss function is formulated as:

\begin{equation}\label{l_rpn}
{
L_{Arpn_{reg}}=-\log{\frac{Intersection + 1}{Union + 1}}
}
\end{equation}

Where $Intersection$ and $Union$ are calculated between the predicted and ground-truth bounding boxes. For $L_{Arpn_{cls}}$, we simply use the cross-entropy loss.

\textbf{LOTM:} To solve the unbalanced problem between the size of background and foreground, we use the class-balanced cross-entropy loss for contour point learning. The loss function is formulated as:

\begin{equation}\label{l_point}
{
L_{cp}=-\frac{N_{neg}}{N} y_i \log{p_i} - \frac{N_{pos}}{N} (1 - y_i) \log(1 - p_i)
}
\end{equation}

Where $y_i$ and $p_i$ denote ground-truth and prediction. $N_{neg}$ and $N_{pos}$ are numbers of negatives and positives respectively. \emph{N} is the sum of $N_{neg}$ and $N_{pos}$. Loss for horizontal prediction $(L_{Hcp})$ and vertical prediction $(L_{Vcp})$ have the identical form as $L_{cp}$.

For $L_{boxclass}$ and $L_{boxreg}$ in box branch , we choose the similar forms in \cite{ren2015faster,girshick2015fast}.

\section{Experiments}

\subsection{Datasets}

\textbf{ICDAR2015} \cite{karatzas2015icdar} is a dataset proposed in the Challenge 4 of ICDAR 2015 Robust Reading Competition. It contains totally 1500 images (1000 training images and 500 testing images) with annotations labeled as 4 vertices at word level. Different from previous datasets containing horizontal texts only, texts in this benchmark have arbitrary orientations.

\textbf{CTW1500} \cite{yuliang2017detecting} is a dataset for curve text detection. It contains 1000 images for training and 500 images for testing. The texts are labeled with 14 boundary points at text-line level.

\textbf{Total-Text} \cite{ch2017total} is a recent challenging dataset. Different from CTW1500, the annotations in this dataset are labelled in word-level. This dataset includes horizontal, multi-oriented, and curved texts. It contains 1255 images for training and 300 images for testing.

\subsection{Implementation Details}

We use the ResNet50 \cite{HeZRS16} pre-trained on ImageNet as our backbone. The model is implemented in Pytorch and trained on 1 NVIDIA TITANX GPU using Adam optimizer \cite{kingma2014adam}. We only use the official training images of each dataset to train our model. Data augmentation includes random rotation, random horizontal flip and random crop. The models are trained 180k iterations in total. Learning rates start from $2.5\times1e-3$, and are multiplied by 0.1 after 120k and 160k iterations.  We use 0.9 momentum and 0.0001weight decay. Multi-scale training is used in our training stage. The short side of images is set to \{400, 600, 720, 1000, 1200\}, and the long side is maintained to 2000. Blurred texts labeled as DO NOT CARE are ignored during training.

As all the datasets use polygon annotations, which are feasible to rebuild texts with arbitrary shapes, we use $distance\_transform\_edt$ in \emph{Scipy} to obtain the two-points wide edge. All the points on the edge are regarded as contour points and used to train our model. The label in Adaptive-RPN can be obtained by using a similar \emph{max-min function} in eq.(\ref{refine}) on ground-truth polygons. During training, we optimize both heatmaps in LOTM with the same supervision.

In testing stage, we use the single scale image as input and evaluate our results through official evaluation protocols. Due to the different scales of test images have a great impact on the detection performance\cite{wang2019arbitrary,liu2018fots}, we scale the images in Total-Text and CTW1500 datasets to $720 \times 1280$, and fix the resolution to $1200\times2000$ for ICDAR 2015. Alpha-Shape Algorithm \cite{akkiraju1995alpha} is used to generate bounding boxes based on contour point candidates.

\subsection{Ablation Study}

We conduct several ablation studies on CTW1500 and Total-Text datasets to verify the effectiveness of Adaptive-RPN and LOTM. All the models are trained using only official training images.

\begin{table}[b!p]
\begin{center}
\begin{tabular}{|l|c|c|c|}
\hline
\emph{n}-points & Recall & Precision & F-measure \\
\hline
5-points & \textbf{85.7} & 81.2 & 83.3 \\
9-points & 84.1 & \textbf{83.7} & \textbf{83.9}\\
\hline
\end{tabular}
\end{center}
\caption{The relationship between performance and the number of pre-defined points used in Adaptive-RPN. 5-points means top-left, top-right, bottom-right, bottom-left and center points.}
\label{tab:number}
\end{table}

\begin{table}[b!p]
\begin{center}
\begin{tabular}{|l|c|c|c|}
\hline
Method & Recall & Precision & F-measure \\
\hline
RPN$\dag$ & 83.8 & 85.1 & 84.5 \\
Adaptive-RPN$\dag$ & \textbf{83.9} & \textbf{86.9} & \textbf{85.4} \\
\hline
RPN* & \textbf{85.6} & 80.8 & 83.1 \\
Adaptive-RPN* & 84.1 & \textbf{83.7} & \textbf{83.9} \\
\hline
& Small & Middle & Large \\
\hline
\textbf{Gain} $\dag$ & 1.4& 0.3& 1.1 \\
\textbf{Gain} * & 1.0& 0.7& 0.8 \\
\hline
\end{tabular}
\end{center}
\caption{The performance gain of Adaptive-RPN. * and $\dag$ are results from CTW1500 and Total-Text respectively. Small, Middle and Large is short for small-size texts, middle-size texts and large-size texts.}
\label{tab:rpn}
\end{table}

\begin{table}[b!p]
\begin{center}
\begin{tabular}{|l|c|c|c|}
\hline
size & Recall & Precision & F-measure \\
\hline
3 & \textbf{83.9} & \textbf{86.9} & \textbf{85.4} \\
5 & 83.6 & 85.7 & 84.7 \\
7 & 83.4 & 85.4 & 84.4 \\
\hline
\end{tabular}
\end{center}
\caption{The relationship between the performance and size of receptive field of texture characteristics in LOTM on Total-Text.}
\label{tab:numberbcd}
\end{table}

\begin{table}[b!p]
\begin{center}
\begin{tabular}{|l|c|c|c|}
\hline
Method & Recall & Precision & F-measure \\
\hline
S-direction & 80.5 & 80.6 & 80.6 \\
Jointly & 82.7 & 85.3 & 84.0 \\
LOTM & \textbf{83.9} & \textbf{86.9} & \textbf{85.4} \\
\hline
\end{tabular}
\end{center}
\caption{The performance gain of LOTM on Total-Text. S-direction means the texture information is only modeled along a single direction (horizontal direction is implemented here). Jointly means the method jointly models the texture information in a $3 \times 3$ convolutional kernel.}
\label{tab:BCDM}
\end{table}

\textbf{Adaptive-RPN:} We first study the relationship between the performance of Adaptive-RPN and number of pre-defined points. As shown in Tab.\ref{tab:number}, Adaptive-RPN implemented with 9 pre-defined points obtains 0.6 \% improvement in F-measure. We set \emph{n} to 9 in the remaining experiments.

To verify the performance gain of the proposed Adaptive-RPN, we conduct several ablation experiments on CTW1500 and Total-Text. LOTM is implemented in all the models. As shown in the top of Tab.\ref{tab:rpn}, Adaptive-RPN obtains 0.9\% and 0.8\% improvement in F-measure on Total-Text and CTW1500 respectively. To further demonstrate the improvement of detecting texts in large variance scale, we further divide the results into three parts according to the size distribution on these two datasets. We consider only the pairs belonging to the same category to be valid for better comparison  (\eg small-size predicted bounding box matches small-size ground-truth bounding box. Note that the number of ignored pairs is almost identical in both methods, which affects little to the results.). As shown in the bottom of Tab.\ref{tab:rpn}, Adaptive-RPN outperforms conventional RPN in F-measure by a large margin in detecting varying-size texts (\eg 1.4\%, 0.3\% and 1.1\% improvement in F-measure in small-size, middle-size and large-size texts respectively on Total-Text).

\textbf{LOTM:} To evaluate the effectiveness of the proposed LOTM, we conduct several experiments on Total-Text. Firstly, we conduct several experiments to study the relationship between the performance and size of convolutional kernels in LOTM. As shown in Tab.\ref{tab:numberbcd}, model implemented with $1\times3$ and $3\times1$ sizes achieves the highest performance (85.4 \% in F-measure). When we further increase the size of receptive field, the performance declines. We infer that the larger receptive field containing more noise is harmful to the performance, which further demonstrates the effectiveness of local texture information modeling. We set the size of convolutional kernels to 3 in the remaining experiments.

Secondly, we evaluate the effectiveness of orthogonal modeling. As shown in Tab.\ref{tab:BCDM}, modeling texture information along only a single direction is a less powerful approach (85.4 \% vs 80.6 \% in F-measure). Compared with jointly modeling the texture information in arbitrary orientations, LOTM obtains a significant improvement in recall, precision and F-measure with 1.2\%, 1.6\% and 1.4\% respectively.

\subsection{Comparisons with State-of-the-Art Methods}

We compare our methods with recent state-of-the-art methods on Total-Text, CTW1500 and ICDAR2015 to demonstrate its effectiveness for arbitrary shape text detection.
\vspace{-1em}
\subsubsection{Evaluation on Curved Text Benchmark}

We evaluate the proposed method on Total-Text to test its performance for curved texts.

\begin{table}[t!p]
\begin{center}
\begin{tabular}{|l|c|c|c|c|c|c|}
\hline
Method & Ext & R & P & F & FPS\\
\hline
SegLink* \cite{shi2017detecting} & - & 23.8 & 30.3 & 26.7 & -\\
EAST* \cite{ZhouYWWZHL17} & - & 36.2 & 50.0 & 42.0 & -\\
Lyu \emph{et al.}\cite{lyu2018mask}& $\checkmark$  & 55.0 & 69.0 & 61.3 & - \\
TextSnake \cite{DBLP:conf/eccv/LongRZHWY18} & $\checkmark$ & 74.5 & 82.7 & 78.4 & -\\
MSR \cite{xue2019msr} & $\checkmark$ & 74.8 & 83.8 & 79.0 &4.3\\
PSENet \cite{Wang_2019_CVPR} & - & 75.1 & 81.8 & 78.3 & 3.9\\
PSENet \cite{Wang_2019_CVPR} & $\checkmark$ & 78.0 & 84.0 & 80.9 & 3.9\\
Wang \emph{et al.}\cite{wang2019arbitrary} & - & 76.2 & 80.9 & 78.5 & -\\
TextDragon \cite{feng2019textdragon} & $\checkmark$ & 74.2 & 84.5 & 79.0 & - \\
TextField \cite{xu2019textfield} & $\checkmark$ & 79.9 & 81.2 & 80.6 & 6\\\
PAN \cite{wang2019efficient}& - & 79.4 & 88.0 & 83.5 & \textbf{39.6}\\
LOMO \cite{zhang2019look} & $\checkmark$ & 75.7 & \textbf{88.6} & 81.6 & 4.4 \\
LOMO$\dag$ \cite{zhang2019look} & $\checkmark$ & 79.3 & 87.6 & 83.3 & - \\
CRAFT \cite{baek2019character} & $\checkmark$ & 79.9 & 87.6 & 83.6 & - \\
\hline
\textbf{Ours} & - & \textbf{83.9} & 86.9 & \textbf{85.4} & 3.8\\
\hline
\end{tabular}
\end{center}
\caption{The single-scale results on Total-Text. * indicates the results from \cite{DBLP:conf/eccv/LongRZHWY18}. Ext is the short for external data used in training stage. $\dag$ means testing at multi-scale setting. The evaluation protocol is DetEval \cite{wolf2006object}.}
\label{tab:total}
\end{table}

\begin{table}[t!p]
\begin{center}
\begin{tabular}{|l|c|c|c|c|c|}
\hline
Method & Ext &R & P & F & FPS\\
\hline
CTPN* \cite{tian2016detecting} & - & 53.8 & 60.4 & 56.9 & 7.1\\
SegLink* \cite{shi2017detecting} & - & 40.0 & 42.3 & 40.8 & 10.7\\
EAST* \cite{ZhouYWWZHL17} & - & 49.1 & 78.7 & 60.4 & 21.2 \\
CTD+TLOC \cite{yuliang2017detecting} & - & 69.8 & 77.4 & 73.4 & 13.3\\
TextSnake \cite{DBLP:conf/eccv/LongRZHWY18} & $\checkmark$& \textbf{85.3} & 67.9 & 75.6 & - \\
PSENet \cite{Wang_2019_CVPR} & - & 75.6 & 80.6 & 78.0 & 3.9\\
PSENet \cite{Wang_2019_CVPR} & $\checkmark$& 79.7 & 84.8 & 82.2 & 3.9\\
Tian \emph{et al.}\cite{tian2019learning} & $\checkmark$& 77.8 & 82.7 & 80.1 & 3\\
Wang \emph{et al.}\cite{wang2019arbitrary} & -& 80.2 & 80.1 & 80.1 & - \\
TextDragon \cite{feng2019textdragon} & $\checkmark$ & 81.0 & 79.5 & 80.2 & - \\
PAN \cite{wang2019efficient}& -& 77.7 & 84.6 & 81.0 & \textbf{39.8}\\
LOMO \cite{zhang2019look} & $\checkmark$& 69.6 & \textbf{89.2 }& 78.4 & 4.4 \\
LOMO$\dag$ \cite{zhang2019look} & $\checkmark$& 76.5 & 85.7 & 80.8 & - \\
CRAFT \cite{baek2019character} & $\checkmark$ & 81.1 & 86.0 & 83.5 & - \\
TextField \cite{xu2019textfield} & $\checkmark$& 79.8 & 83.0 & 81.4 & 6 \\
MSR \cite{xue2019msr} & $\checkmark$& 78.3 & 85.0 & 81.5 & 4.3\\
\hline
\textbf{Ours} & -& 84.1 & 83.7 & \textbf{83.9} & 4.5\\
\hline
\end{tabular}
\end{center}
\caption{The single-scale results on CTW1500. * indicates the results from \cite{yuliang2017detecting}. Ext is the short for external data used in training stage. $\dag$  means testing at multi-scale setting.}
\label{tab:ctw}
\vspace{-1em}
\end{table}

As shown in Tab.\ref{tab:total}, with the help of Adaptive-RPN and false-positive suppression, the proposed method achieves a new state-of-the-art result of 83.9\%, 86.9\% and 85.4\% in recall, precision and F-measure respectively without external data, and outperforms existing state-of-the-art methods (\eg LOMO \cite{zhang2019look}, PAN \cite{wang2019efficient},  PSE\cite{Wang_2019_CVPR}) by a large margin. Meanwhile, it also achieves impressive speed (3.8 FPS). Though CRAFT \cite{baek2019character} use additional character-level annotations to train their model, our method trained with only original annotations outperforms CRAFT \cite{baek2019character} by 1.8 \% in F-measure. Besides, LOMO \cite{zhang2019look} uses external images to train their model and further tests their results at multi-scale level. Our method, which is trained with only official data and tested at single scale, outperforms LOMO \cite{zhang2019look} by 2.1\% in F-measure. The visualization of curved text detection results are shown in Fig.\ref{result}(a).

\begin{table}[t!p]
\begin{center}
\begin{tabular}{|l|c|c|c|c|c|}
\hline
Method & Ext & R & P & F & FPS\\
\hline
EAST \cite{ZhouYWWZHL17} & $\checkmark$  & 73.5 & 83.6 & 78.2 & 13.2\\
Liao et al. \cite{LiaoZSXB18} & $\checkmark$  & 79.0 & 85.6 & 82.2 & 6.5\\
Lyu et al. \cite{LyuYWYB18} & $\textbf{}\checkmark$ & 70.7 & \textbf{94.1} & 80.7 & 3.6\\
FOTS \cite{liu2018fots} & $\checkmark$ & 82.0 & 88.8 & 85.3 & 7.8 \\
PixelLink \cite{DBLP:conf/aaai/DengLLC18} & - & 81.7 & 82.9 & 82.3 & 7.3\\
MSR \cite{xue2019msr} & $\checkmark$ & 78.4 & 86.6 & 82.3 & 4.3\\
PSENet \cite{Wang_2019_CVPR} & - & 79.7 & 81.5 & 80.6 & 1.6\\
PSENet \cite{Wang_2019_CVPR} & $\checkmark$ & 84.5 & 86.9 & 85.7 & 1.6 \\
PAN \cite{wang2019efficient}& - & 77.8 & 82.9 & 80.3 & \textbf{26.1}\\
TextDragon \cite{feng2019textdragon} & $\checkmark$ & 81.8 & 84.8 & 83.1 & - \\
LOMO \cite{zhang2019look} & $\checkmark$ & 83.5 & 91.3 & 87.2 & 3.4 \\
TextField* \cite{xu2019textfield} & $\checkmark$ & 83.9 & 84.3 & 84.1 & 1.8 \\
Liu \emph{et al.} \cite{liu2019omnidirectional} & $\checkmark$ & 83.8 & 89.4 & 86.5 & - \\
Tian \emph{et al.} \cite{tian2019learning} & $\checkmark$ & 85.0 & 88.3 & 86.6 & 3\\
CRAFT \cite{baek2019character} & $\checkmark$ & 84.3 & 89.8 & 86.9 & - \\
Wang \emph{et al.} \cite{wang2019arbitrary} & - & 83.3 & 90.4 & 86.8 & - \\
Wang \emph{et al.}$\dag$ \cite{wang2019arbitrary} & - & 86.0 & 89.2 & \textbf{87.6 }& - \\
\hline
\textbf{Ours} & - & \textbf{86.1} & 87.6 & 86.9 & 3.5 \\
\hline
\end{tabular}
\end{center}
\caption{The single-scale results on ICDAR2015. * means testing at multi-scale setting. $\dag$ means SE blocks \cite{hu2018squeeze} implemented in their backbone.}
\label{tab:ic15}
\vspace{-1em}
\end{table}

\subsubsection{Evaluation on Long Curved Text Benchmark}

To show the performance of our ContourNet for long curved texts, we compare its performance with state-of-the-arts on CTW1500 dataset, which is annotated at text-line level.

\begin{figure*}[t!p]
\centering
\subfigure[Total-Text]{
\begin{minipage}[b]{0.95\linewidth}
\includegraphics[width=1\linewidth]{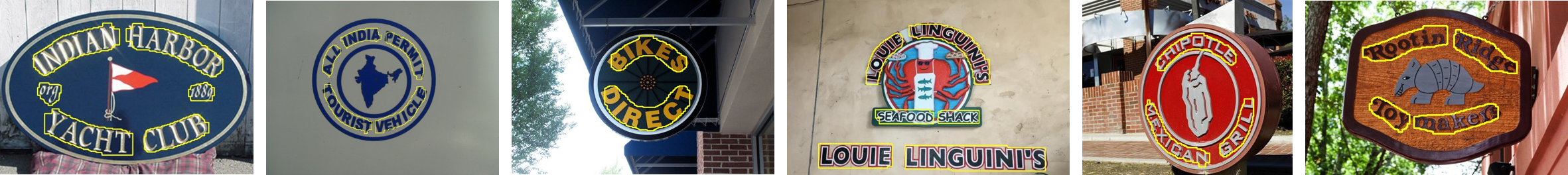}\\
\vspace{-1.5em}
\end{minipage}}
\vspace{-1em}
\subfigure[CTW1500]{
\begin{minipage}[b]{0.95\linewidth}
\includegraphics[width=1\linewidth]{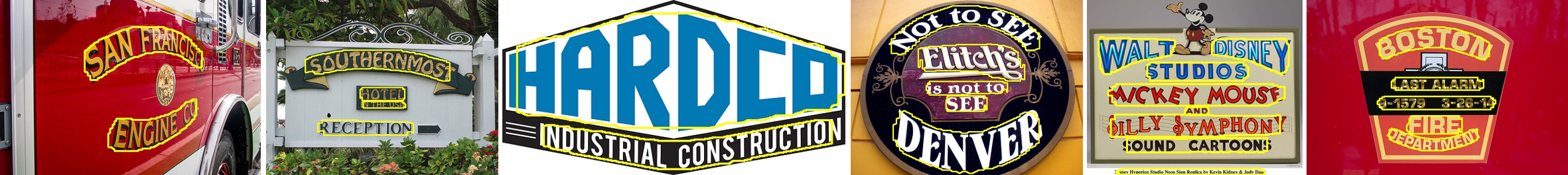}\\
\vspace{-1.em}
\end{minipage}}
\subfigure[ICDAR2015]{
\begin{minipage}[b]{0.95\linewidth}
\includegraphics[width=1\linewidth]{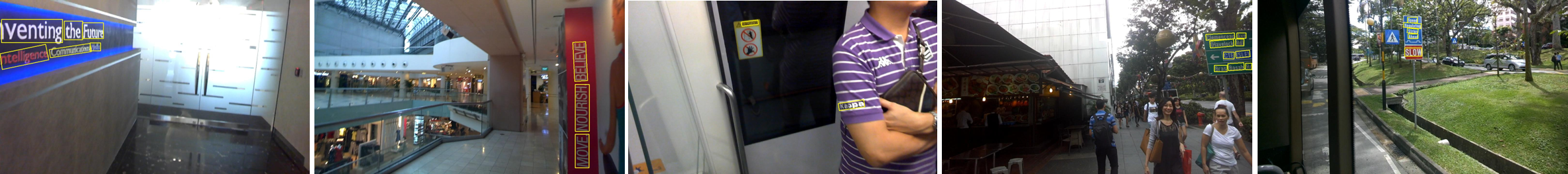}\\
\vspace{-1.5em}
\end{minipage}}
\caption{Results on different datasets. (a) results on Total-Text; (b) results on CTW1500; (c) results on ICDAR2015.}\label{result}
\vspace{-1em}
\end{figure*}

As shown in Tab.\ref{tab:ctw}, the proposed method is much better than other counterparts including CTD+TLOC \cite{yuliang2017detecting}, MSR \cite{xue2019msr}, TextSnake \cite{DBLP:conf/eccv/LongRZHWY18}, which are designed for curved texts. Though text region refinement in LOMO \cite{zhang2019look} achieves promising results on representing long texts, our ContourNet, which benefits from Adaptive-RPN, achieves much higher performance (83.9 \% vs 80.8\% in F-measure). Compared with MSR\cite{xue2019msr} which also uses contour points to describe text regions, ours shows advantages in both recall and F-measure without external data for training, where the relative improvement reaches 5.8\% and 2.4\% respectively. In addition, the proposed method runs at 4.5 FPS on this dataset, which is faster than most recent methods. The visualization of long curved text detection results are shown in Fig.\ref{result}(b).

\subsubsection{Evaluation on Multi-oriented Text Benchmark}

We evaluate our method on ICDAR 2015 to test its performance for multi-oriented texts. RoIAlign \cite{he2017mask} is used for the generation of proposal features on this dataset.

Several experimental results are shown in Tab.\ref{tab:ic15}. Our method achieves 86.9 \% in F-measure, which is only a little lower than Wang \emph{et al.}\cite{wang2019arbitrary} (87.6\% in F-measure). However, they implement Squeeze-and-Excitation (SE) blocks \cite{hu2018squeeze} in their backbone, which is more powerful to recalibrate channel-wise feature responses. When implemented without SE blocks, their method achieves 86.8 \% in F-measure, which is lower than our method. The visualization of multi-oriented text detection results are shown in Fig.\ref{result}(c).

\subsection{Effectiveness of ContourNet}

We further demonstrate the effectiveness of our method in the following two aspects. More discussions about this part are shown in the supplementary.

\textbf{Effectiveness of Adaptive-RPN.}
As the large scale variance problem exists in scene text detection, conventional RPN obtains a coarse localization of text region when the regression distance is large or the target box has quite different ratio to default box. Benefiting from the awareness of shape information and the scale-invariant training object, the proposed Adaptive-RPN performs better in these cases and achieves finer localization of text regions. Some qualitative examples of conventional RPN and proposed Adaptive-RPN are shown in the supplementary.

\textbf{Effectiveness of false-positive suppression.}
\textbf{1) Quantification}. The value of $\theta$ in Point Re-scoring Algorithm affects the ratio of suppressed FPs to caused false negetives (FNs). The value of ratio is considerable when $\theta$ goes from 0.1 to 0.9 (an elaborated chart is shown in the supplementary). Thus, our method is much more effective in suppressing FPs than in causing FNs. \textbf{2) Qualitative analysis.} Though few FNs are caused, it is worth mentioning that the retained positive points with strong texture information in both orthogonal directions are able to accurately represent texts (see in Fig.\ref{recent}). \textbf{3)} Implemented with conventional RPN, our method can achieve 84.5\% and 83.1\% in F-measure on Total-Text and CTW1500 respectively, surpassing most methods in Tab.\ref{tab:total} and Tab.\ref{tab:ctw}. Though it is hard to verify which representation is better for arbitrary-shaped text detection (\eg region predictions \cite{wang2019arbitrary,wang2019efficient}, contour points \cite{xue2019msr}, adaptive points \cite{wang2019arbitrary}, etc.), FP problem is the uniform challenge in each method. In this regard, our method obtains a significant improvement compared with the previous.

\vspace{-0.5em}
\section{Conclusion}

In this paper, we propose a novel scene text detection method (ContourNet) to handle the false positives in text representation and  the large scale variance problem. ContourNet mainly consists of three parts including Adaptive-RPN, LOTM and \emph{Point Re-scoring Algorithm}. Adaptive-RPN localizes the preliminary proposals of texts by bounding the spatial extend of several semantic points. LOTM models the local texture information in two orthogonal directions and represents text region with contour points. \emph{Point Re-scoring Algorithm} filters FPs by considering the response values in both orthogonal directions simultaneously. The effectiveness of our approach has been demonstrated on several public benchmarks including long, curved and oriented text cases. In future works, we prefer to develop an end-to-end text reading system.

\vspace{-0.5em}
\section*{Acknowledgments}

This work is supported by the National Key Research and Development Program of China (2017YFC0820600), the National Nature Science Foundation of China (61525206, U1936210), the Youth Innovation Promotion Association Chinese Academy of Sciences (2017209), the Fundamental Research Funds for the Central Universities under Grant WK2100100030.

{\small
\bibliographystyle{ieee_fullname}
\bibliography{egbib}
}

\end{document}